\pdfoutput=1

\documentclass[11pt]{article}

\usepackage[final]{acl}

\usepackage{times}
\usepackage{latexsym}

\usepackage[T1]{fontenc}

\usepackage[utf8]{inputenc}

\usepackage{microtype}

\usepackage{inconsolata}

\usepackage{graphicx}
\usepackage{kotex}
\usepackage{amsmath}
\usepackage{booktabs}
\usepackage[capitalize]{cleveref}
\crefname{section}{Sec.}{Secs.}
\Crefname{section}{Section}{Sections}
\Crefname{table}{Table}{Tables}
\crefname{table}{Tab.}{Tabs.}
\usepackage{amssymb}
\usepackage{comment}
\usepackage{subcaption}
\usepackage{multirow}
\usepackage{multicol}

\title{PruneCD: Contrasting Pruned Self Model to Improve Decoding Factuality}

\author{
  Byeongho Yu$^{1}$$^{*}$ \quad 
  Changhun Lee$^{2}$$^{*}$ \quad 
  Jungyu Jin$^{3}$
  \quad Eunhyeok Park$^{3}$\\[0.3em]
  $^{1}$Department of Computer Science and Engineering \\
  $^{2}$Department of Convergence IT Engineering \\
  $^{3}$Graduate School of Artificial Intelligence \\
  Pohang University of Science and Technology (POSTECH) \\[0.3em]
  \texttt{\{bhyu418, changhun.lee, jgjin0317, eh.park\}@postech.ac.kr}
}

\begin{document}
\maketitle
\renewcommand{\thefootnote}{\fnsymbol{footnote}}
\footnotetext[1]{ These authors contributed equally.}
\renewcommand{\thefootnote}{\arabic{footnote}}
\begin{abstract}

To mitigate the hallucination problem in large language models, DoLa exploits early exit logits from the same model as a contrastive prior. However, we found that these early exit logits tend to be flat, low in magnitude, and fail to reflect meaningful contrasts. To address this, we propose PruneCD, a novel contrastive decoding method that constructs the amateur model via layer pruning rather than early exit. This design leads to more informative and well-aligned logits, enabling more effective contrastive decoding. Through qualitative and quantitative analyses, we demonstrate that PruneCD consistently improves factuality with minimal inference overhead, offering a robust and practical approach to mitigating hallucinations in LLMs.

\end{abstract}

\section{Introduction}

Contrastive Decoding (CD)~\cite{li2023contrastive}, though originally proposed to improve decoding diversity, has recently emerged as a promising approach to mitigating the hallucination problem in large language models (LLMs)~\cite{grattafiori2024llama, yang2025qwen3}, a phenomenon where autoregressive, probability-based sampling yields fluent yet factually incorrect outputs, especially for questions that are underrepresented or unseen during training.
By contrasting the confidence levels of expert and amateur models for the same input, CD selectively promotes responses where the mature model is confident while the less reliable model is uncertain, thereby producing more trustworthy and grounded outputs. Following its promising results in related studies~\cite{o2023contrastive, shi2024trusting}, CD has attracted growing interest.

\begin{figure}[t]
    \centering    \includegraphics[width=1.0\linewidth]{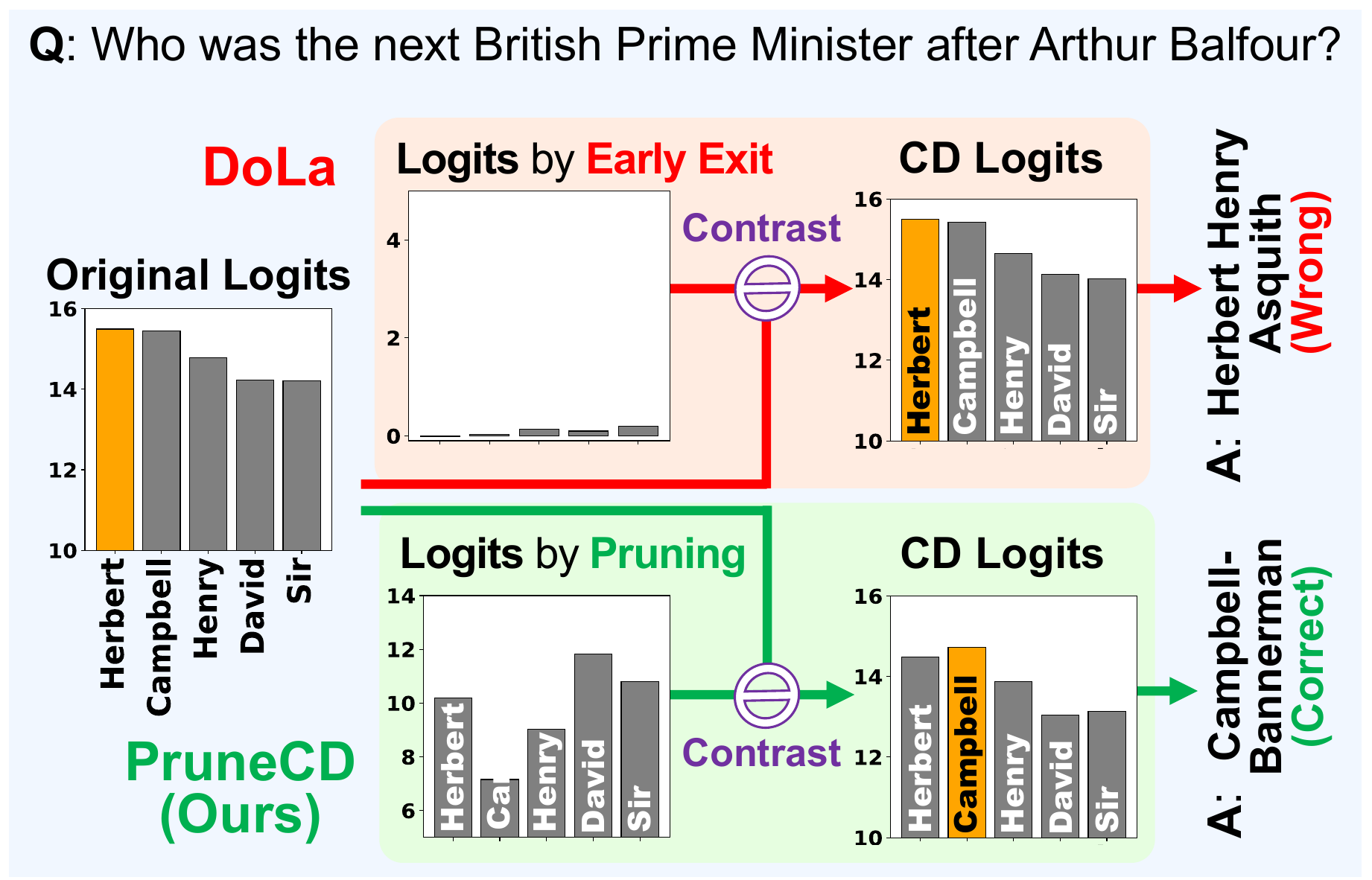}\vspace{-2mm}
    \caption{Comparison of amateur model logits based on early exit (\textcolor{red}{Red} region) and layer pruning (\textcolor{green}{Green} region), along with the resulting CD logits from each approach. 
    }
    \vspace{-5mm}
    \label{fig:logits}
\end{figure}

While early CD methods relied on two separate models to create contrasting confidence levels, recent approaches achieve this contrast within a single model. A representative example is DoLa \cite{chuang2023dola}, which introduces early exit to extract intermediate logits as less confident predictions. 
Leveraging premature early exit logits as a contrastive prior to strengthen factual grounding in deeper layers has demonstrated more reliable generation in LLMs.
However, we found that amateur logits produced by an early exit mechanism are typically flat and uninformative, 
delivering only marginal gains in contrastive decoding and thereby limiting their effectiveness.

To address this, we propose \textbf{PruneCD}. Rather than relying on early exit, PruneCD constructs the amateur through fine-grained layer pruning, producing more informative contrasts to enhance factuality, as illustrated in \Cref{fig:logits}. While retaining the key benefit of enabling CD without an additional model, PruneCD combines insights from prior work to offer a more intuitive interpretation and practical implementation. 
Our qualitative results demonstrate its robustness and superior performance across diverse model scales and tasks.
Our implementation is available at \url{https://github.com/hoeng4/PruneCD}.

\section{Related work}
\paragraph{Contrastive Decoding}
The strong potential of Contrastive Decoding (CD) has sparked a wave of important follow-up studies~\cite{o2023contrastive, shi2024trusting}. Unlike these prior approaches, our proposed PruneCD achieves superior factual generation while preserving the key advantage of DoLa~\cite{chuang2023dola}, enabling contrastive decoding without requiring a separately trained amateur model or additional fine-tuning.

\noindent\textbf{Advanced Decoding Methods} \quad
In addition to contrastive decoding, recent methods such as Activation Decoding~\cite{chen2024context} and END~\cite{wu2025improve} enhance the reliability of LLMs by guiding decoding through activation entropy and cross-layer entropy, respectively. We include comparisons with these strong baselines to demonstrate the superior performance of our proposed method. Further details are provided in \Cref{sec:ablation_entropy_baselines}.


\section{Motivation}

\begin{figure}
  \centering
  \begin{subfigure}{0.48\linewidth}
    \centering
    \includegraphics[width=\linewidth]{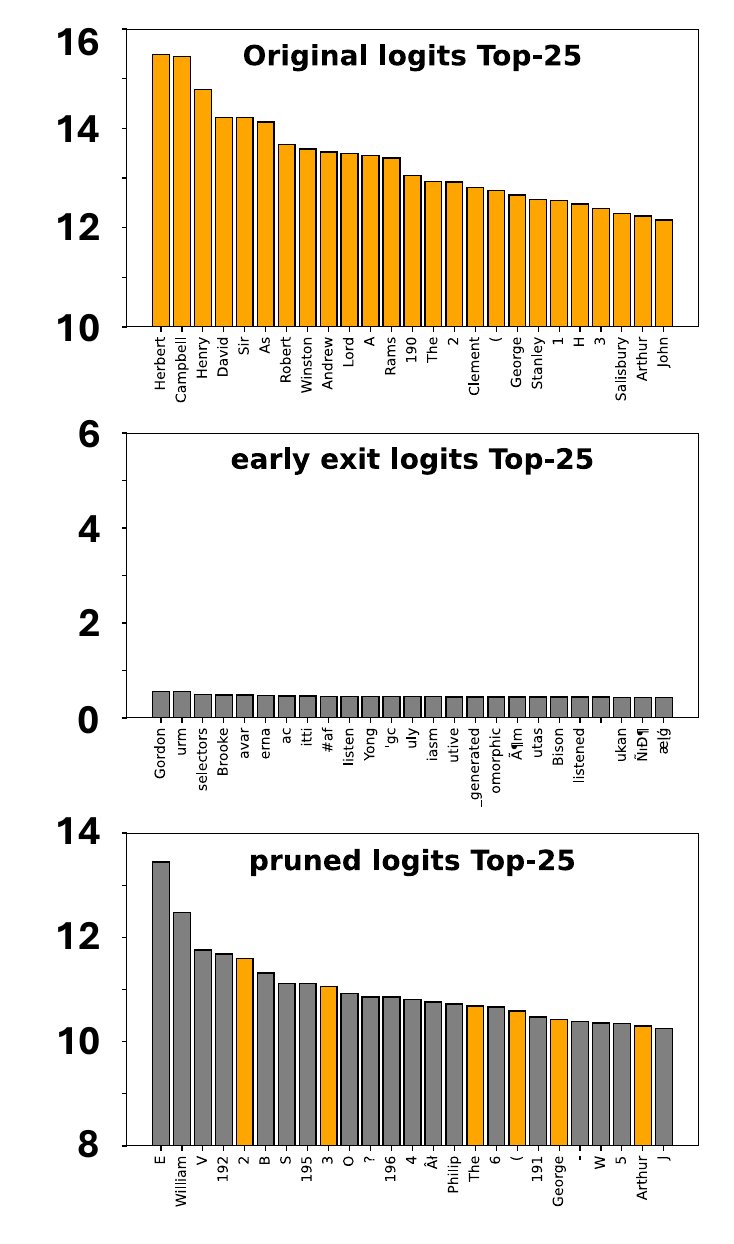} 
    \caption{Logits}
    \label{fig:topk_logits}
  \end{subfigure}
  \begin{subfigure}{0.49\linewidth}
    \centering
    \includegraphics[width=\linewidth]{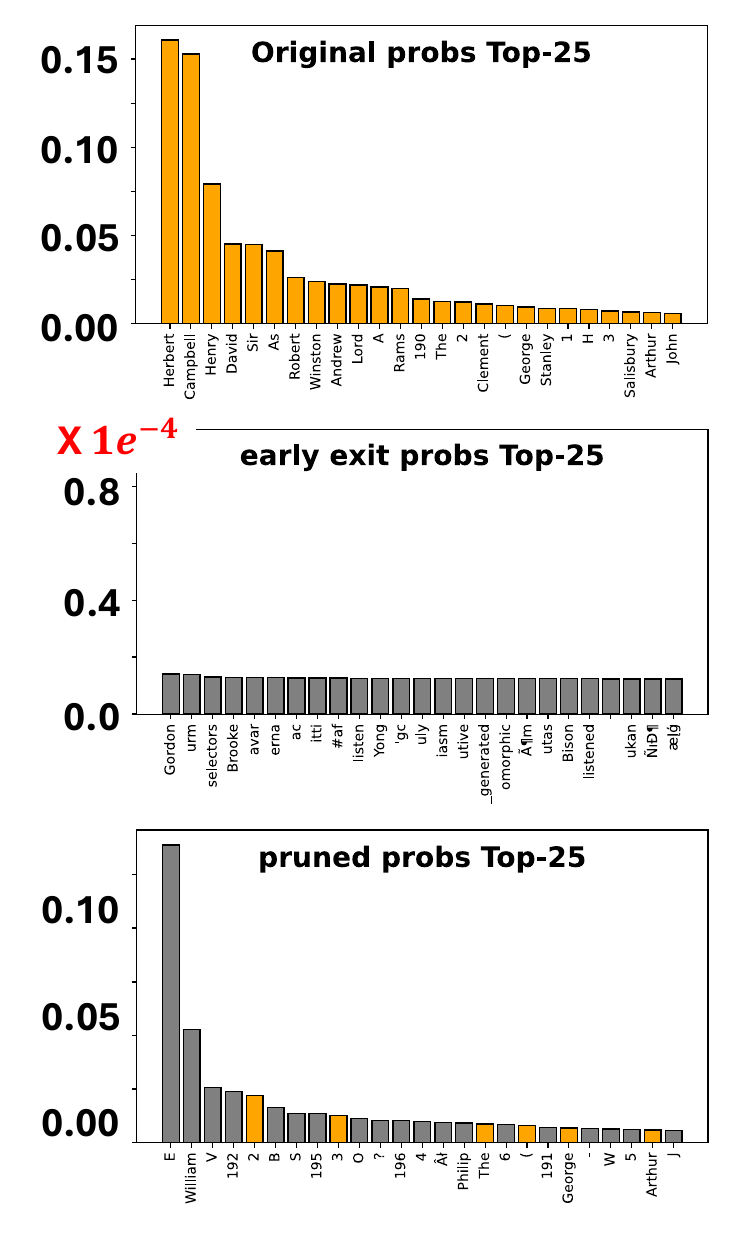}
    \caption{Probabilities}
    \label{fig:topk_probs}
  \end{subfigure}
  \caption{Top-25 tokens of (i) the original, (ii) early exit, and (iii) layer-pruned versions of the model, presented as (a) logits and (b) soft-max probabilities, each sorted in descending order. The orange bar indicates that the token is included in the original top-25 set.}
  \label{fig:figure2}
  \vspace{-5mm}
\end{figure}

DoLa~\cite{chuang2023dola} is a strong baseline that enables contrastive decoding without the need for a separate amateur by leveraging early exit outputs from the expert model as amateur logits. This approach is motivated by two key insights: (1) factual information tends to be injected in the higher layers, and (2) the Jensen-Shannon divergence between logits effectively identifies this injection point.

However, this intuition does not consistently hold in practice. For instance, in multilingual reasoning tasks, \citet{zhu2024multilingual} reported that the expert and amateur logits in DoLa may represent different language domains, weakening the contrast. Our analysis further reveals that DoLa often selects the earliest possible exit layer, thereby terminating prematurely (see \Cref{sec:appendix_DoLa}).

As shown in \Cref{fig:logits}, the resulting early exit logits are flat and low in magnitude, deviating significantly from those of the expert model. This weak contrast provides little influence during contrastive decoding and fails to correct hallucinated responses from greedy decoding. 

This observation can be explained by circuit-level analyses of transformer models, which suggest distinct functional roles across layers:
(1) different layers serve distinct functions~\cite{ferrando2024primer}, and
(2) the upper transformer layers amplify probabilities shaped earlier in the computation~\cite{lieberum2023does, yu2023neuron}.
From this perspective, early exit misses the final sharpening stage, making it a suboptimal choice for contrastive decoding.

\setlength{\abovedisplayskip}{6pt}
\setlength{\belowdisplayskip}{6pt}

\paragraph{Formal Definition} To move beyond intuition and analyze this issue, we introduce two properties of amateur logits: \textbf{flatness} and \textbf{informativeness}. Flatness, measured by entropy, reflects how diffuse the probability distribution is: high entropy implies an almost uniform and unconfident distribution. Informativeness measures how well the amateur logits align with the expert logits, quantified by the overlap between their top-k tokens. The amateur logits should be non-flat and informative, not collapsing to uniformity but retaining meaningful structure relative to the expert logits.

Formally, let \(z^{(e)}, z^{(a)} \in \mathbb{R}^V\) be expert and amateur logits with soft-max distributions \(p^{(e)}, p^{(a)}\), respectively. Eq.~\ref{eq:flatness} and Eq.~\ref{eq:informativeness} are the definition of flatness and informativeness.
\begin{align}
  H(p) &= -\sum_i p_i \log p_i \,,
  \label{eq:flatness} \\ 
  O_k &= \bigl\lvert \mathrm{Top}_k(z^{(e)}) \cap 
                  \mathrm{Top}_k(z^{(a)}) \bigr\rvert.
  \label{eq:informativeness}
\end{align}

\begin{figure*}[t]
    \centering
    \includegraphics[width=\linewidth]{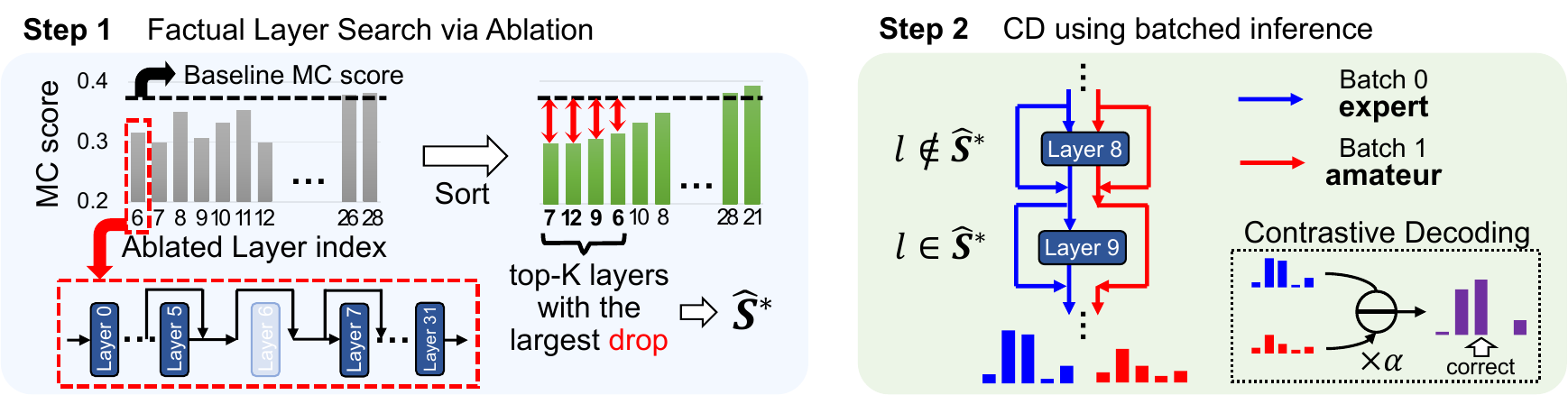}\vspace{-2mm}
    \caption{Overall pipeline of the proposed PruneCD. In this example, layer 6, 7, 9, and 12 are pruning set, $\hat{S}^*$.}
    \vspace{-5mm}
    \label{fig:pipeline}
\end{figure*}

\paragraph{Empirical Analysis} With the definitions of flatness and informativeness, we illustrate these properties using the same example question shown in \Cref{fig:logits}. First, we evaluated early exit (DoLa) on Llama-3.1-8B-Instruct. In \Cref{fig:topk_logits}, remarkably, \textbf{none} of the top-25 tokens from the early exit logits overlap with the expert’s top-25 predictions (all shown in gray), and their values fall below 0.3, indicating a flat and misaligned distribution. After softmax normalization (\Cref{fig:topk_probs}), \textbf{the early exit distribution collapses to an almost uniform} probability of $10^{-5}$ across the tokens, offering virtually no meaningful guidance for contrastive decoding.

In contrast, we explore constructing the amateur model through layer pruning rather than early exit. Rather than terminating at an early exit point, we prune eight intermediate layers following the exit point and resume computation from the subsequent layers. We refer to the logits produced by this modified path as pruned logits. As shown in \Cref{fig:figure2}, the pruned logits retain several overlapping tokens with the original output and maintain comparable magnitude. More importantly, they assign differentiated, and thus informative, probability mass across those tokens, providing a meaningful signal.

These observations suggest that the following inequalities are expected to hold:
\begin{align}
H(p^{(a)}_\text{early exit}) &\gg H(p^{(e)}), \\
O_k(z^{(a)}_\text{early exit}, z^{(e)}) &\ll O_k(z^{(a)}_\text{pruned}, z^{(e)}).
\end{align}

\begin{table}[t]
\centering
\small
\begin{tabular}{c c c c}
\toprule
\textbf{} & \textbf{full model} & \textbf{early exit} & \textbf{pruned} \\
\midrule
Entropy  & 1.3717 & 11.7498 & 2.2669 \\
\bottomrule
\end{tabular}

\begin{tabular}{c c c}
\toprule
\textbf{} & \textbf{early exit} & \textbf{pruned} \\
\midrule
Average Overlapping \quad  & 0.4343 & 15.4978 \\
\bottomrule
\end{tabular}

\caption{\textbf{(Top)} Average Entropy of full model, early exit, and pruned logits and \textbf{(Bottom)} the average number of overlapping tokens among the top-25 predictions between full model logits and early exit/pruned logits.}
\vspace{-4mm}
\label{tab:entropy_overlap}
\end{table}

To verify that these relationships are not limited to a single case, we computed flatness
and informativeness on 1,000 TriviaQA prompts using Llama-3.1-8B-Instruct (\Cref{tab:entropy_overlap}). The average entropy of early exit logits from DoLa was 11.75, compared to 1.37 for the full model logits, confirming that early exit yields significantly flatter distributions. The entropy of pruned logits was 2.27, much closer to the expert, indicating sharper and more confident outputs. Similarly, the average top-25 token overlap between early exit and expert logits was only 0.43, while pruning achieved an overlap of 15.50, showing substantially better alignment.

Taken together, these analyses suggest that layer pruning provides amateur logits that are degraded yet still informative, forming a more reliable basis for contrastive decoding.

\section{PruneCD} Building on this insight, we propose \textbf{PruneCD}, a carefully designed algorithm for pruning-target selection that maximizes the effectiveness of CD. \Cref{fig:pipeline} outlines the overall pipeline of PruneCD. 
 
\subsection{Layer pruning based amateur model}
Given a sequence of tokens $\{x_1, x_2, \ldots, x_{t-1}\}$, the CD score for the next token $x_t$ is defined as follows:
\begin{equation}
    \label{eq:cd}
    \begin{split}
      &CD_{\text{score}}(x_t; x_{<t}) \\
      &= \log p^{(e)}(x_t \mid x_{<t})  - \lambda \log p^{(a)}(x_t \mid x_{<t}),
    \end{split}
\end{equation}
where $\lambda$ is the CD temperature to control the strength of contrasting, and $p^{(e)}$ and $p^{(a)}$ are probabilities of the expert and amateur models, respectively. We define the \textbf{expert model} as the model using full $n$ decoder layer stacks \( \mathcal{L} = \{ L_0, L_1, \dots, L_{n-1} \} \),
and the \textbf{amateur model} as a model using partial layers that prunes a subset \( \mathcal{S} \subset \mathcal{L} \), resulting in the model with layer set \(\mathcal{L} \setminus \mathcal{S} \).

We define \( f(x_{<t}; \mathcal{L}') \) as the output probability distribution over the next token 
obtained by forwarding the input sequence through the specified decoder layers set \( \mathcal{L}' \), followed by an unembedding layer and softmax. 
Accordingly, the probabilities from the expert and amateur models are defined as:
\begin{align}
    \label{eq:probability}
    p^{(e)}(x_t \mid x_{<t}) &= f(x_{<t}; \mathcal{L}), \\
    p^{(a)}(x_t \mid x_{<t}) &= f(x_{<t}; \mathcal{L} \setminus \mathcal{S}).
\end{align}
We describe the method for selecting the pruning set $S$ in the following subsection.

\begin{table*}[t]
\centering
\small
\setlength{\tabcolsep}{5pt}
\renewcommand{\arraystretch}{0.9}
\begin{tabular}{c c ccc ccc cc cc c}
\toprule
\textbf{Llama}
& \multirow{2}{*}{\textbf{Method}}
& \multicolumn{3}{c}{\textbf{TruthfulQA Gen}} 
& \multicolumn{3}{c}{\textbf{TruthfulQA MC}} 
& \multicolumn{2}{c}{\textbf{TriviaQA}} 
& \multicolumn{2}{c}{\textbf{NQ}}
& \textbf{StrQA} \\
\cmidrule(lr{2pt}){3-5} \cmidrule(lr{2pt}){6-8} \cmidrule(lr){9-10} \cmidrule(lr){11-12} \cmidrule(lr){13-13}
\addlinespace[2pt]
\textbf{Model} & & \%Truth & \%Info & \%T*I & MC1 & MC2 & MC3 & EM & F1 & EM & F1 & \%Acc \\
\midrule
                    & Greedy     & 88.86 & 42.03 & 37.34 & 38.61 & 58.63 & 30.17 & 67.00 & 66.27 & 36.98 & 34.90 & 75.41 \\
                    &  DoLa      & 79.11 & 66.46 & 52.58 & 38.73 & 56.66 & 27.66 & 67.29 & 65.42 & 37.34 & 33.63 & 73.41 \\
3.1-8B-Inst         &  ActD      & 83.67 & 54.30 & 45.44 & 36.33 & 55.46 & 27.53 & 67.44 & 66.52 & 37.37 & 35.20 & 75.68 \\
                    &  END       & 86.71 & 47.72 & 41.38 & 40.00 & 60.28 & \textbf{32.08} & 67.11 & 66.26 & 36.90 & 34.71 & 75.33 \\ 
                    &  \textbf{PruneCD}      &  92.78 & 85.19 & \textbf{79.04} & \textbf{42.78} & \textbf{61.65} & 31.65 & \textbf{67.49} & \textbf{66.53} & \textbf{37.62} & \textbf{35.42} & \textbf{76.55} \\
\midrule
                    & Greedy     & 84.81 & 44.68 & 37.90 & 31.01 & 52.08 & 25.13 & 52.30 & 51.82 & 30.83 & 28.48 & 66.99 \\
                    &  DoLa      & 72.03 & 74.18 & 53.43 & 33.16 & 51.74 & 23.97 & 52.28 & 51.81 & 31.02 & 28.57 & 68.47 \\
3.2-3B-Inst         &  ActD      & 74.18 & 74.05 & 54.93 & 33.54 & 54.44 & 26.02 & 53.37 & 52.78 & 31.22 & 28.90 & 68.69 \\
                    &  END       & 87.34 & 40.13 & 35.05 & 32.15 & 52.43 & 26.06 & 52.35 & 51.92 & 30.86 & 28.50 & 67.07 \\
                    &  \textbf{PruneCD}    & 91.39 & 65.70 & \textbf{60.04} & \textbf{36.08} & \textbf{56.39} & \textbf{28.02} & \textbf{53.39} & \textbf{52.96} & \textbf{31.55} & \textbf{29.20} & \textbf{69.87} \\
\midrule
                    & Greedy     & 73.42 & 51.27 & 37.64 & 26.84 & 45.92 & 21.63 & 33.37 & 33.57 & 18.81 & 17.33 & 59.21 \\
                    &  DoLa      & 57.34 & 89.24 & 51.17 & 27.34 & 43.70 & 21.24 & 33.40 & 33.67 & 18.75 & 17.19 & 60.39 \\
3.2-1B-Inst         &  ActD      & 58.10 & 78.10 & 45.38 & 26.96 & \textbf{50.38} & 22.28 & 33.49 & 33.77 & 19.00 & 17.48 & 61.48 \\
                    &  END       & 68.99 & 63.92 & 44.10 & 26.84 & 46.99 & 21.29 & 33.55 & 33.88 & 18.86 & 17.21 & 60.79 \\
                    &  \textbf{PruneCD}      &  66.46 & 87.47 & \textbf{58.13} & \textbf{27.59} & 46.83 & \textbf{22.49} & \textbf{34.21} & \textbf{34.35} & \textbf{19.36} & \textbf{17.97} & \textbf{61.70} \\

\bottomrule
\end{tabular} \vspace{-2mm}
\caption{Comparison of performance across multiple Llama family models and datasets. We \textbf{bold} the best score.}
\vspace{-4mm}
\label{tab:multi_dataset_metrics}
\end{table*}

\subsection{Factual Layer Search via Ablation}
\label{sec:factual_layer_search}

In PruneCD, we need to identify the layers that are most dominantly responsible for encoding factual knowledge. Therefore, we systematically ablate combinations of layers and measure their impact on factuality, and then select the layer set that results in the greatest factuality score drop.

Given $n$ layer stacks \( \mathcal{L} = \{ L_0, L_1, \dots, L_{n-1} \} \), we can search optimal pruning set $S^*$ by:
\begin{equation}
    \label{eq:search}
S^* = \underset{\substack{S \subseteq \mathcal{L}, \:\: |S| < k}}{\arg\max} ( \mathrm{MC}(\mathcal{L}) - \mathrm{MC}(\mathcal{L} \setminus S) ),
\end{equation}
where MC($\mathcal{L}'$) denotes the multiple-choice score of a model that utilizes only the subset of layers $\mathcal{L}'$, and $k$ is the maximum number of layers to be pruned.
Specifically, we use the validation set of the TruthfulQA \cite{lin2022truthfulqa} and adopt the MC1 score for assessment.
However, the search space comprises at least ${}_n \mathrm{C}_k$ combinations, making exhaustive search infeasible (e.g., ${}_{32} \mathrm{C}_4 = 35,960$).

To overcome this, we instead ablate one layer at a time and select the top-$k$ layers that lead to the greatest degradation in MC score (\Cref{fig:pipeline} Left). 
We then use these $k$ layers as the pruning set ${\hat{S}}^*$.
This approach requires only $n$ evaluations and is significantly more efficient. We can further reduce search space to sub-$n$ by additional filtering, which details are provided in the \Cref{sec:ablation_ppl_filter}.

Since the searched set ${\hat{S}}^*$ leads to substantial degradation in truthfulness when pruned, CD with the corresponding amateur model is expected to yield more factual decoding results. 


\subsection{Efficient Inference}
\label{sec:efficient_inference}
Since $\hat{S}^*$ can be determined statically in advance, both amateur and expert probabilities can be computed simultaneously with a single forward pass. As shown on the Right side of \Cref{fig:pipeline}, we utilized batched inference to minimize the overhead. For the single sample, PruneCD assigns the full forward computation to batch 0 and applies skip connections for the layers in $\hat{S}^*$ in batch 1.

\section{Experiments}

\paragraph{Models}
We evaluate a range of recent models of varying sizes: Llama-3.1-8B-Instruct \cite{grattafiori2024llama}, Llama-3.2-3B-Instruct, and Llama-3.2-1B-Instruct. 
These models have strong general capabilities while being lightweight in size.

\noindent\textbf{Datasets} \quad
We use six benchmark datasets: TruthfulQA \cite{lin2022truthfulqa}, StrategyQA (StrQA) \cite{geva2021did}, TriviaQA \cite{joshi2017triviaqa}, and Natural Questions (NQ) \cite{kwiatkowski2019natural} for evaluating factuality, and GSM8K \cite{cobbe2021training} and VicunaQA \cite{vicuna2023} to evaluate different domains. Further details are provided in \Cref{sec:appendix_datasets}.
 
\noindent\textbf{Baselines} \quad
We include Greedy Decoding and DoLa. 
We also adopt Activation Decoding (ActD) \cite{chen2024context} and END \cite{wu2025improve}, two advanced decoding methods specifically designed to improve factuality.

\noindent\textbf{Implementation Details} \quad
We provide the full implementation details in \Cref{sec:appendix_implementation}, including the adaptive plausibility constraint \cite{li2023contrastive} and the repetition penalty, as in DoLa. PruneCD has two hyperparameters: 1) CD temperature $\lambda \in \mathbb{R}$, 2) the number of pruned layers $k \in \mathbb{Z}$. For fair comparison, our hyperparameters and those of other baselines were all searched through a validation run on the respective benchmark separately. Details of the hyperparameters for our method and the baselines are provided in \Cref{sec:ablation_hyperparameters}.

\subsection{Overall Results}
The overall results are presented in \Cref{tab:multi_dataset_metrics}.
Across all evaluated models, our method outperformed all baselines (DoLa, ActD, and END) on nearly all tasks.
In the open-ended generation setting of TruthfulQA, DoLa showed a decrease in Truthfulness compared to Greedy decoding, whereas our method improved not only Informativeness but also Truthfulness. 
Notably, PruneCD achieved consistently superior performance to the baselines on TriviaQA, NQ, and StrQA.

On smaller models such as Llama-3.2-3B-Instruct, our method demonstrated particularly large performance gains over both Greedy decoding and DoLa.
This highlights the effectiveness of our approach in addressing hallucination in the context of the recent trend toward deploying smaller models.
Furthermore, our method outperforms baselines even on the 1B model, demonstrating strong generalization ability and robustness across both tasks and model sizes. 

\begin{table}[t]
\centering
\small
\begin{tabular}{c c c c}
\toprule
\textbf{Method} & \textbf{3.1-8B-Inst} & \textbf{3.2-3B-Inst} & \textbf{3.2-1B-Inst} \\
\midrule
Greedy  & 77.18 \hphantom{\textcolor{green!50!black}{+0.99}} 
        & 66.03 \hphantom{\textcolor{green!50!black}{+2.81}} 
        & 35.86 \hphantom{\textcolor{green!50!black}{+3.18}} \\
DoLa    & 78.17 \textcolor{green!50!black}{+0.99} 
        & 66.41 \textcolor{green!50!black}{+0.38} 
        & 36.39 \textcolor{green!50!black}{+0.53} \\
ActD    & 78.47 \textcolor{green!50!black}{+1.29} 
        & 68.84 \textcolor{green!50!black}{+2.81} 
        & 37.60 \textcolor{green!50!black}{+1.74} \\
END     & 77.63 \textcolor{green!50!black}{+0.45} 
        & 61.79 \textcolor{red}{–4.24} 
        & 35.86 \textcolor{black!50}{+0.00} \\
\textbf{PruneCD} & \textbf{81.43} \textcolor{green!50!black}{\textbf{+4.25}} 
        & \textbf{70.58} \textcolor{green!50!black}{\textbf{+4.55}} 
        & \textbf{39.04} \textcolor{green!50!black}{\textbf{+3.18}} \\
\bottomrule
\end{tabular} \vspace{-2mm}
\caption{GSM8K accuracy across Llama family models.}
\label{tab:GSM8K}
\end{table}

\begin{table}[t]
\centering
\small
\begin{tabular}{c c c c}
\toprule
\textbf{Method} & Wins & Ties & Losses \\
\midrule
\textcolor{green!50!black}{PruneCD} vs Greedy  & \textcolor{green!50!black}{\textbf{111}} & 21 & 28 \\
\textcolor{green!50!black}{PruneCD} vs DoLa    & \textcolor{green!50!black}{\textbf{83}} & 38 & 39 \\
\bottomrule
\end{tabular} \vspace{-2mm}
\caption{
Pairwise comparison results on VicunaQA. 
}
\vspace{-4mm}
\label{tab:vicunaqa}
\end{table}

\begin{table}[t]
\centering
\small
\begin{tabular}{c c c c}
\toprule
 & Greedy & DoLa & PruneCD \\
\midrule

Inference speed & 35.8 & 30.8 & 33.7 \\

\bottomrule
\end{tabular} 

\vspace{-2mm}
\caption{
TruthfulQA Gen inference speed (token/s).
}
\vspace{-4mm}
\label{tab:inference_latency}
\end{table}

\subsection{Evaluation on different domains}

To further assess generalizability, we also ran experiments on GSM8K and VicunaQA. Unlike factoid QA benchmarks, these datasets emphasize reasoning and instruction-following, offering a broader evaluation of decoding strategies.

On GSM8K, which requires multi-step reasoning, PruneCD consistently outperformed all baselines across different Llama models (\Cref{tab:GSM8K}). The green numbers in the table denote improvements over greedy decoding, showing gains of roughly 3–5 pp across model scales. These gains highlight the method’s ability to generalize beyond factual QA into arithmetic and logical domains.

For VicunaQA, we used Llama-3.1-8B-Instruct as the model and GPT-4o \cite{gpt4o}\footnote{We used the model version \texttt{gpt-4o-2024-08-06}.} as the judge, applying reversed answer order to control for position bias. As shown in Table~\ref{tab:vicunaqa}, PruneCD demonstrated consistently higher win rates than both greedy decoding and DoLa across all pairwise matchups, confirming its effectiveness in generating high-quality, instruction-following responses.

\subsection{Inference efficiency}
\label{sec:latency_results}
We measure the inference latency of Greedy decoding, DoLa, and PruneCD on the TruthfulQA open-ended generation task using an A100 80GB GPU.
The resulting decoding speeds (tokens/s) are summarized in \Cref{tab:inference_latency}, where PruneCD achieves comparable throughput to Greedy decoding.

\subsection{Additional analyses}
\label{sec:appendix_results}
We also include additional analyses in the Appendix. These cover fixed-hyperparameter evaluations showing that PruneCD is robust without task-specific tuning (\Cref{sec:fixed_hyper_results}), experiments on Mistral-7B-Instruct-v0.3 confirming improvements beyond the Llama family (\Cref{sec:mistral_result}), and qualitative examples illustrating corrections of common misconceptions (\Cref{sec:appendix_examples}).

Multilingual Contrastive Decoding (MCD) \cite{zhu2024multilingual} focuses on the specific setting of non-English multilingual reasoning and therefore differs fundamentally from our approach in both target objectives and pruning set selection strategy. However, for completeness, we include a comparison with MCD in \Cref{sec:appendix_multilingual}.

\section{Conclusion}
Hallucination remains a major challenge in LLMs, prompting active research into solutions such as contrastive decoding (CD). In this paper, we introduced \textbf{PruneCD}, a novel CD method that improves factuality by constructing the amateur model via layer pruning, offering a more principled approach to address the limitations of early exit.
Comprehensive experiments across multiple QA datasets and varying sizes of models demonstrate that PruneCD consistently outperforms existing baselines, with inference latency comparable to greedy decoding. 
As a simple yet effective solution, we believe that PruneCD will serve as an important milestone for future research on hallucination mitigation.

\newpage
\section*{Limitations}
As described in \Cref{sec:efficient_inference} and \Cref{sec:latency_results}, our batched inference implementation of PruneCD achieves generation latency comparable to greedy decoding.
However, a potential limitation arises from the use of batched inference: it may increase memory footprint due to additional internal activations.
While this memory overhead could become noticeable when using very large batch sizes, we did not observe any significant overhead under typical small-batch settings, such as those used in our evaluation environment.

Moreover, during the factual layer search, more advanced approaches could be explored. For example, gradient-based strategies might help identify informative pruning targets more effectively. The search granularity could also be refined beyond the decoder layer level, such as operating at the attention or feed-forward block level. We leave these directions as promising avenues for future work.


\section*{Acknowledgments}
This work was supported by IITP and NRF grant funded by the Korea government(MSIT) (No. RS-2019-II191906, RS-2024-00415602, RS-2023-00228970, RS-2025-02218259).



\newpage
\appendix

\section{Advanced Decoding methods}
\label{sec:ablation_entropy_baselines}
Activation Decoding \cite{chen2024context} identified that the activation entropy across intermediate layers tends to be lower for the correct answer token and proposed an entropy-based probability adjustment accordingly.
END \cite{wu2025improve} found that the prediction probability of factuality tokens tends to grow sharply along layers, and suggests controlling decoding using cross-layer entropy.

\section{Experiment Settings in Figures}
\label{sec:figure_settings}
\begin{figure}
    \centering
    \includegraphics[width=1.0\linewidth]{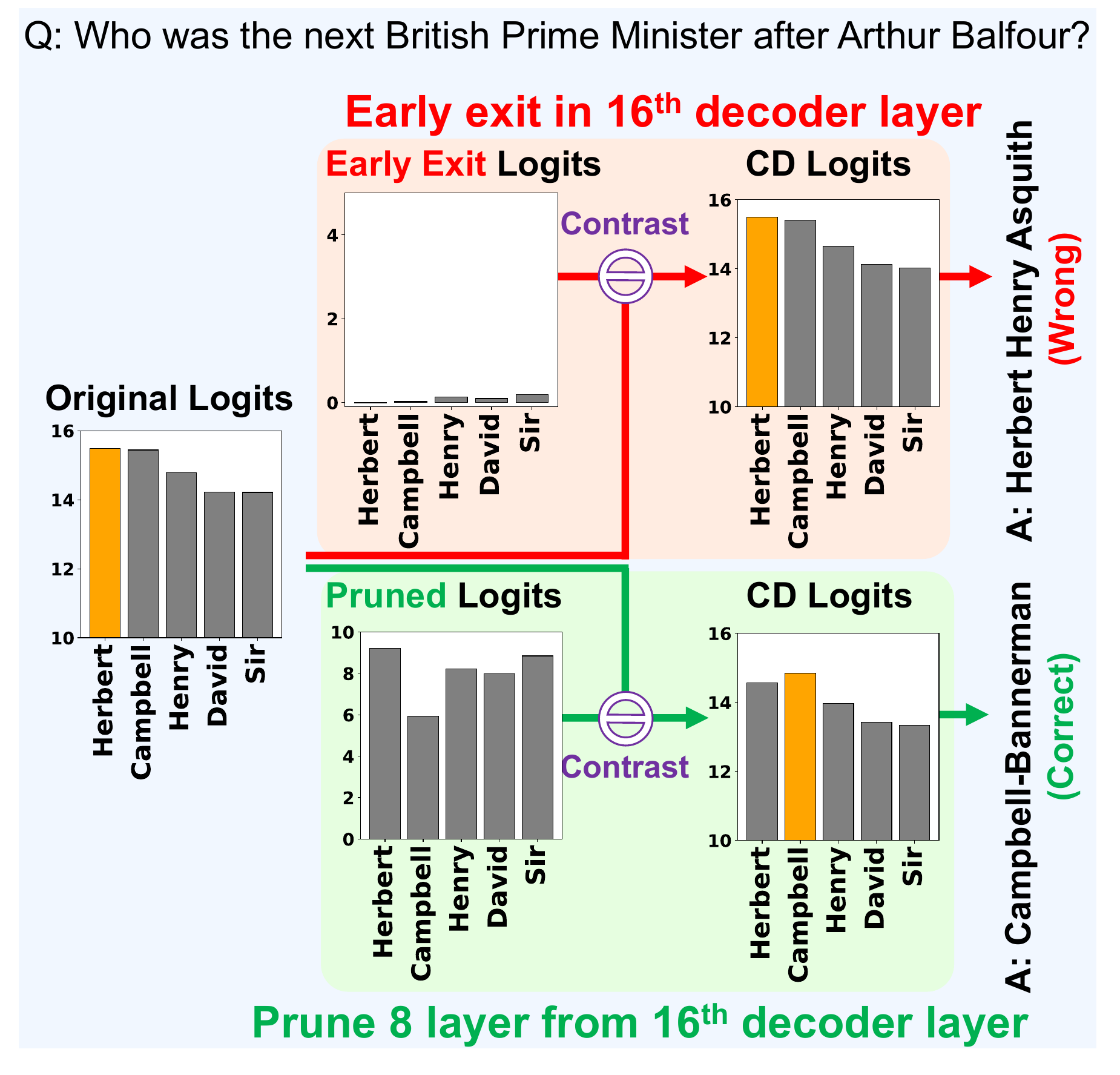}
    \caption{Toy example corresponding to \Cref{fig:figure2}, illustrating that layer pruning produces amateur logits better suited for contrastive decoding than early exit. 
    The resulting pruned logits exhibit a noticeably richer than the early exit logits, confirming the information advantage of layer pruning.
}
    \vspace{-4mm}
    \label{fig:toy_example}
\end{figure}

All panels in \Cref{fig:logits}, \Cref{fig:figure2}, and \Cref{fig:toy_example} are produced with Llama-3.1-8B-Instruct evaluated on TriviaQA in a few-shot setting.
For the DoLa, the early exit layer is chosen in upper-half bucket.
In \Cref{fig:logits}, the pruning set consists of layer 6, 7, 9, and 12, and CD temperature is set to 0.1. For \Cref{fig:toy_example}, CD temperature is set to 0.1.

\section{Experimental Details}
\label{sec:appendix}
\subsection{Datasets}
\label{sec:appendix_datasets}
We first consider \textbf{TruthfulQA}, with 790 QA samples, a widely used benchmark for evaluating the truthfulness and informativeness of LLMs, which consists of two task formats: multiple choice and open-ended generation.
The multiple-choice setting is evaluated using the MC1, MC2, and MC3 metrics. For open-ended generation, prior works relied on a fine-tuned GPT-3 model for evaluation, which is no longer supported. Therefore, we instead use publicly available fine-tuned Llama-2-7B\footnote{allenai/truthfulqa-truth-judge-llama2-7B \\ allenai/truthfulqa-info-judge-llama2-7B} models for evaluation. To evaluate general knowledge extraction quality, we include widely used question answering datasets such as \textbf{TriviaQA} and \textbf{Natural Questions}. These are evaluated using Exact Match (EM) and F1 scores. We use validation set of TriviaQA (11313 QA samples) and Natural Questions (3610 QA samples). \textbf{StrategyQA} is used to assess the model’s chain-of-thought reasoning abilities.

\textbf{GSM8K} is used to measure mathematical reasoning ability, consisting of diverse grade-school math word problems. We follow standard evaluation using exact match accuracy on the test set (1319 problems). \textbf{VicunaQA} is employed to assess instruction-following and open-ended response quality. Following prior work, evaluation is performed with an LLM judge (GPT-4o) comparing model outputs on 80 diverse questions across categories such as writing, roleplay, coding, and reasoning.

\subsection{Full Implementation Details}
\label{sec:appendix_implementation}
The original CD paper \cite{li2023contrastive} used the plausibility constraint:
\begin{align}
    &\mathcal{V}_{\text{head}}(x_{<i}) = \\
    &\big\{ x_i \in \mathcal{V} \!:\!\!\!\!\!\! \nonumber \!\quad p_{\text{EXP}}(x_i \mid x_{<i}) 
    \geq \alpha \max_w p_{\text{EXP}}(w \mid x_{<i}) \big\}
    \label{eq:plausibility_constraint}
\end{align}
This constraint ensures that only plausible candidates from the expert model are considered for contrastive selection. As in CD paper and other baselines, we apply the same constraint in our method to maintain compatibility and avoid selecting implausible or low-confidence tokens, and as in CD paper, we fixed the value of $\alpha = 0.1$.
We also apply a repetition penalty $\theta=1.2$, following DoLa and other baseline methods.

\paragraph{Implementation Details for Baselines} For DoLa, we used the official implementation from HuggingFace's Transformers library to perform evaluation on generation tasks. For Activation Decoding and Multilingual Contrastive Decoding (MCD) \cite{zhu2024multilingual}, we conducted evaluation using the official code released on GitHub.
Since END did not have publicly available code, we implemented it ourselves for reproduction.

\begin{table*}[t]
\centering
\small
\begin{tabular}{c c cc cc c c}
\toprule
\multirow{2}{*}{\textbf{Llama}}
& \multirow{2}{*}{\textbf{Method}}
& \multicolumn{2}{c}{\textbf{TriviaQA}} 
& \multicolumn{2}{c}{\textbf{NQ}}
& \textbf{StrQA}
& \textbf{GSM8K} \\
\cmidrule(lr){3-4} \cmidrule(lr){5-6} \cmidrule(lr){7-7} \cmidrule(lr){8-8}
\textbf{Model} & & EM & F1 & EM & F1 & \%Acc & \%Acc\\
\midrule
3.1-8B-Inst & Greedy & 67.0 & 66.3 & 37.0 & 34.9 & 75.4 & 77.18 \\
            & DoLa   & 67.3 & \textcolor{red}{65.4} & \textbf{37.3} & \textcolor{red}{33.6} & \textcolor{red}{73.4} & 78.17 \\
            & ActD   & 67.3 & \textbf{66.6} & 37.1 & \textcolor{red}{34.9} & \textcolor{red}{74.7} & 78.47\\
            & \textbf{PruneCD}   & \textbf{67.5} & 66.4 & \textbf{37.3} & \textbf{35.2} & \textbf{75.6} & \textbf{81.43}\\
\midrule
3.2-3B-Inst & Greedy     & 52.3 & 51.8 & 30.8 & 28.5 & 67.0 & 66.03 \\
&DoLa            & \textcolor{red}{52.3} & \textcolor{red}{51.8} & 30.9 & 28.7 & 68.5 & \textcolor{red}{64.67} \\
&ActD            & 52.4 & 51.9 & \textcolor{red}{30.8} & \textcolor{red}{28.5} & 68.7 & \textcolor{red}{65.88} \\
&\textbf{PruneCD}   & \textbf{53.1} & \textbf{52.7} & \textbf{31.5} & \textbf{29.2} & \textbf{69.3} & \textbf{69.52} \\
\midrule
3.2-1B-Inst & Greedy     & 33.4 & 33.6 & 18.8 & 17.3 & 59.2 & 35.86 \\
& DoLa            & \textcolor{red}{33.4} & 33.7 & \textcolor{red}{18.3} & \textcolor{red}{17.0} & 60.4 & \textcolor{red}{35.71} \\
& ActD            & 33.5 & 33.8 & \textbf{19.0} & 17.5 & \textbf{61.5} & 37.60 \\
& \textbf{PruneCD}   & \textbf{33.9} & \textbf{34.1} & \textbf{19.0} & \textbf{17.8} & \textbf{61.5} & \textbf{37.91} \\

\bottomrule
\end{tabular} \vspace{-2mm}
\caption{Performance comparison across multiple tasks with fixed hyperparameter settings in the Llama family}
\vspace{-4mm}
\label{tab:fixed_hyperparameter}
\end{table*}

\subsection{Hyperparameters}
\label{sec:ablation_hyperparameters}
\paragraph{PruneCD}
We consider two hyperparameters in our method: (1) the number of pruned layers, $k \in \mathbb{Z}$, and (2) the CD temperature, $\lambda \in \mathbb{R}$.
The number of pruned layers is selected differently depending on the model size. Specifically, for Llama-3.1-8B-Instruct with 32 layers, we use $k \in [3, 8]$; for Llama-3.2-3B-Instruct with 28 layers, $k \in [3, 6]$; and for Llama-3.2-1B-Instruct with 16 layers, $k \in [1, 4]$.
For the CD temperature $\lambda$, we use values in $[1.0, 1.5]$ for the TruthfulQA dataset, and values in $[0.1, 0.3]$ for the other QA tasks.

\paragraph{DoLa}
DoLa introduces a hyperparameter in the form of a layer bucket, which defines the candidate set of premature layers for early exit. For models with 40 or fewer layers, two variants are considered: DoLa-Lower, which uses the lower half of the decoder layers (early layers) as candidate premature layers, and DoLa-Upper, which uses the upper half (later layers). We perform a bucket-wise search between these two configurations and report results using the better-performing bucket.
\paragraph{Activation Decoding}
Activation Decoding (ActD) involves two hyperparameters: (1) the information layer $l$ used for activation calculation, and (2) the scaling factor $\lambda$ applied when adjusting the next-token probability distribution based on entropy. For Llama-3.1-8B-Instruct, we consider $\{26, 28, 30, 32\}$ as candidate info layers. For Llama-3.2-3B-Instruct with 28 layers, we use $\{22, 24, 26, 28\}$, and for Llama-3.2-1B-Instruct with 16 layers, we use $\{12, 14, 16\}$. This follows the original paper’s approach of selecting different candidates of the intermediate layers based on model size.
For the entropy scaling factor $\lambda$, we perform a hyperparameter search over the range $\{0.2, 0.4, 0.6, 0.8\}$. \textit{Note that this $\lambda$ is unrelated to the $\lambda$ used as the CD temperature in PruneCD.}
\paragraph{END}
END involves two hyperparameters: (1) the probability threshold $\alpha$, introduced to improve the efficiency of entropy computation, and (2) the scaling factor $\lambda$, used to adjust the next-token probability distribution based on entropy. The role of $\alpha$ corresponds to the hyperparameter used in the plausibility constraint, as described in \Cref{sec:appendix_implementation}. While PruneCD and all other baselines compared in this paper fix $\alpha = 0.1$, for END we follow the original paper and perform a search over the set \{0.1, 0.01, 0.001\}. Although this search has little effect on open-ended generation tasks, it contributes to notable performance improvements in multiple-choice tasks. Similarly, for the scaling factor $\lambda$, we follow the paper and search over \{1.0, 2.0, 3.0\} for open-ended generation tasks (e.g., TruthfulQA-Gen, StrategyQA), and \{0.25, 0.375, 0.5\} for multiple-choice and QA datasets. \textit{Note that this $\lambda$ is unrelated to the $\lambda$ used as the CD temperature in PruneCD.}

\section{Fixed Hyperparameter Results}
\label{sec:fixed_hyper_results}

To evaluate the robustness of decoding methods without task-specific hyperparameter tuning, we conducted additional experiments using a fixed hyperparameter configuration across all benchmarks (\Cref{tab:fixed_hyperparameter}). Specifically, the number of pruned layers in PruneCD was selected solely based on validation performance on TruthfulQA-MC1, and the CD temperature $\lambda$ was fixed at 0.2 for all tasks. For DoLa, we applied the best bucket configuration from TruthfulQA-MC1. For Activation Decoding, we selected informative layers in the same way, but allowed to tune its temperature on a per-task basis, giving it a relative advantage in this comparison.

\begin{table*}[t]
\centering
\small
\begin{tabular}{c ccc ccc cc cc c c}
\toprule
\multirow{2}{*}{\textbf{Method}}
& \multicolumn{3}{c}{\textbf{TruthfulQA Gen}} 
& \multicolumn{3}{c}{\textbf{TruthfulQA MC}} 
& \multicolumn{2}{c}{\textbf{TriviaQA}} 
& \multicolumn{2}{c}{\textbf{NQ}}
& \textbf{StrQA} 
& \textbf{GSM8k} \\
\cmidrule(lr{2pt}){2-4} \cmidrule(lr{2pt}){5-7} \cmidrule(lr){8-9} \cmidrule(lr){10-11} \cmidrule(lr){12-12} \cmidrule(lr){13-13}
\addlinespace[2pt]
 & \%Truth & \%Info & \%T*I & MC1 & MC2 & MC3 & EM & F1 & EM & F1 & \%Acc & \%Acc \\
\midrule
Greedy     & 78.23 & 76.46 & 59.81 & 47.09 & \textbf{65.13} & \textbf{35.99} & 65.62 & 64.90 & 34.76 & 33.21 & 73.71 & 51.55 \\
DoLa      & 78.10 & 91.14 & 71.18 & 44.30 & 63.63 & 32.02 & 65.46 & 64.41 & 35.04 & 32.46 & 70.96 & 50.49\\
ActD      & 77.09 & 88.23 & 68.01 & 43.54 & 53.00 & 31.82 & 65.84 & 65.05 & 35.10 & 33.25 & 73.97 & 52.31\\
END       & 77.97 & 90.63 & 70.67 & 46.46 & 64.00 & 36.31 & 65.57 & 64.79 & 35.01 & 33.19 & 71.66 &51.40\\ 
\textbf{PruneCD}      &  85.06 & 90.89 & \textbf{77.31} & \textbf{48.10} & 60.38 & 35.70 & \textbf{66.61} & \textbf{65.85} & \textbf{35.46} & \textbf{34.00} & \textbf{74.28} & \textbf{54.13}\\
\bottomrule
\end{tabular} \vspace{-2mm}
\caption{Performance comparison across multiple tasks with Mistral-7B-Instruct-v0.3}
\vspace{-4mm}
\label{tab:mistral}
\end{table*}

\Cref{tab:fixed_hyperparameter} summarizes these results: bold values mark the best method, while red numbers indicate cases where performance does not improve over greedy decoding.
Even under these constraints, PruneCD consistently outperformed greedy decoding across all tasks, whereas DoLa and Activation Decoding often failed to yield gains and sometimes degraded performance.
This demonstrates the robustness and generality of PruneCD, which remains effective with a single hyperparameter setting across tasks of diverse formats and reasoning demands—a property especially valuable for real-world deployment where exhaustive per-task hyperparameter tuning is infeasible.

\section{Results on the Mistral model family}
\label{sec:mistral_result}

To assess generality beyond the Llama family, we further evaluated PruneCD on Mistral-7B-Instruct-v0.3. As shown in \Cref{tab:mistral}, PruneCD achieved strong performance across QA tasks, consistently matching or surpassing other decoding baselines. Notably, on GSM8K, PruneCD improved accuracy by +2.58 points over greedy decoding (51.55\% to 54.13\%), demonstrating its robustness on multi-step numerical reasoning. These results confirm that PruneCD generalizes well beyond Llama and remains effective across diverse model families.

\section{Early Exit locations in DoLa}
\label{sec:appendix_DoLa}
\begin{figure}[t]
  \centering
  \begin{subfigure}{0.48\linewidth}
    \centering
    \includegraphics[width=\linewidth]{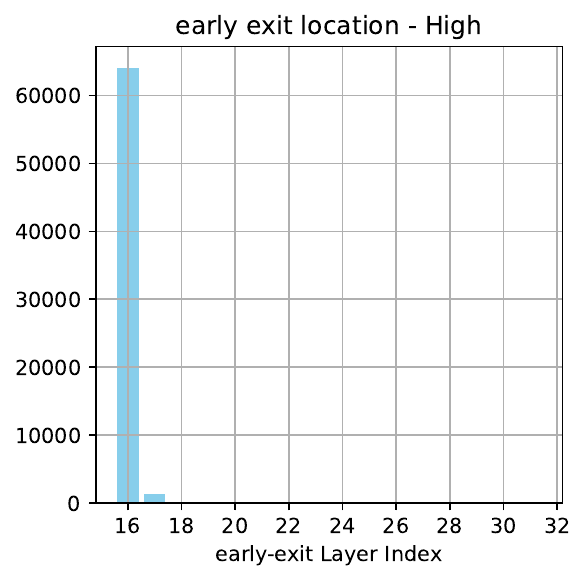}
    \caption{Upper-half bucket}
    \label{fig:dola_high}
  \end{subfigure}\hfill
  \begin{subfigure}{0.48\linewidth}
    \centering
    \includegraphics[width=\linewidth]{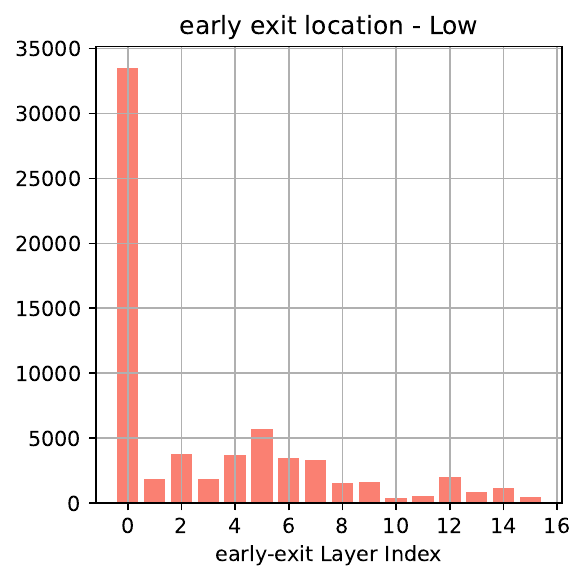}
    \caption{Lower-half bucket}
    \label{fig:dola_low}
  \end{subfigure}
  \vspace{-2mm}
  \caption{Distribution of the contrasting layer selected by DoLa on Llama-3.1-8B-Instruct for the TruthfulQA-MC task when the candidate set is the upper half or lower half of the decoder layers.}
  \label{fig:early_exit_location}
  \vspace{-4mm}
\end{figure}

DoLa first partitions the decoder into fixed buckets and then selects, within each chosen bucket, the layer whose logits maximize the Jensen–Shannon divergence (JSD) from the final logits. \Cref{fig:early_exit_location} illustrates the distribution of the selected contrasting-layer indices for Llama-3.1-8B-Instruct under two different bucket configurations: (1) \textbf{Upper-half bucket}, spanning layers 16–31, and (2) \textbf{Lower-half bucket}, covering layers 0–15. In the upper-half bucket setting, the selected layer consistently collapses to layer \textbf{16}; in the lower-half setting, it collapses to layer \textbf{0}. This demonstrates that when the candidate set is a contiguous block, DoLa’s JSD-based selection criterion overwhelmingly favors the lower boundary layer of the bucket, offering virtually no variation in the choice of contrasting layers. 

\begin{figure*}[t]
    \centering
    \includegraphics[width=0.8\linewidth]{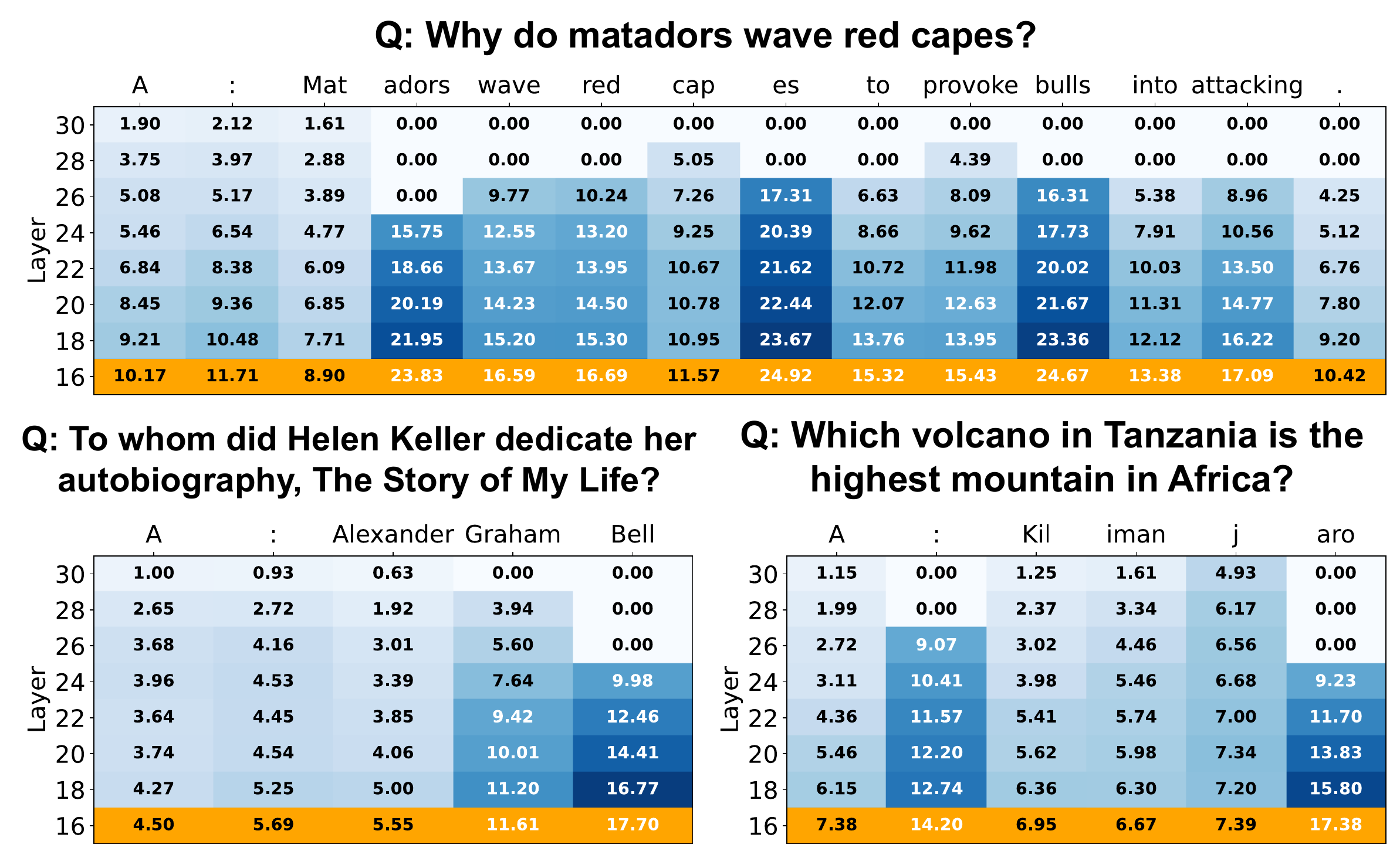}
    \vspace{-2mm}
    \caption{For each generated token, the figure shows the Jensen–Shannon divergence between the early exit logits at every decoder layer and the final logits.  Orange cells indicate, for each token, the layer that attains the largest divergence; these layers coincide with the contrasting layers chosen by DoLa.}
    \label{fig:jsd}
\end{figure*}

Furthermore, we provide several examples from different datasets to visualize how the JSD values evolve across layers and how the selected contrasting layer is determined under the upper-half bucket configuration in \Cref{fig:jsd}. In particular, for sample inputs from the TruthfulQA and TriviaQA datasets, we observe that layer 16 is consistently chosen as the contrasting layer, regardless of whether the token in question is related to factualness or informativeness.

\begin{table*}[t]
\centering
\small
\setlength{\tabcolsep}{4pt}
\renewcommand{\arraystretch}{0.9}
\begin{tabular}{c ccc ccc cc cc c}
\toprule

\textbf{Method} 
& \multicolumn{3}{c}{\textbf{TruthfulQA Gen}} 
& \multicolumn{3}{c}{\textbf{TruthfulQA MC}} 
& \multicolumn{2}{c}{\textbf{TriviaQA}} 
& \multicolumn{2}{c}{\textbf{NQ}}
& \textbf{StrQA} \\
\cmidrule(lr){2-4} \cmidrule(lr){5-7} \cmidrule(lr){8-9} \cmidrule(lr){10-11} \cmidrule(lr){12-12}
& \%T & \%I & \%T*I &  MC1 & MC2 & MC3 & EM & F1 & EM & F1 & \%Acc \\
\midrule
    Greedy              & 88.86 & 42.03 & 37.34 & 38.61 & 58.63 & 30.17 & 67.00 & 66.27 & 36.98 & 34.90 & 75.41 \\
    \textbf{PruneCD}                & 92.78 & 85.19 & 79.04 & 42.78 & 61.65 & 31.65 & 67.49 & 66.53 & 37.62 & 35.42 & 76.55 \\
    \textbf{PruneCD w/o filtering} & 87.97 & 86.58 & 76.17 & 43.04 & 62.34 & 31.84 & 67.31 & 66.49 & 37.67 & 35.43 & 76.16 \\
\bottomrule
\end{tabular} \vspace{-2mm}
\caption{Performance comparison with and without our perplexity-based filtering in Llama-3.1-8B-Instruct model.}
\vspace{-4mm}
\label{tab:barrier_no_barrier}
\end{table*}

\section{Perplexity-based search space filtering}
\subsection{Details of the method}
\label{sec:ablation_ppl_filter}
In \Cref{sec:factual_layer_search}, we first perform perplexity-based filtering to narrow down the set of candidate layers, and then conduct multiple choice score based ablation on the remaining candidates. The detailed procedure is as follows.

\begin{figure*}[t]
    \centering
    \includegraphics[width=1.0\linewidth]{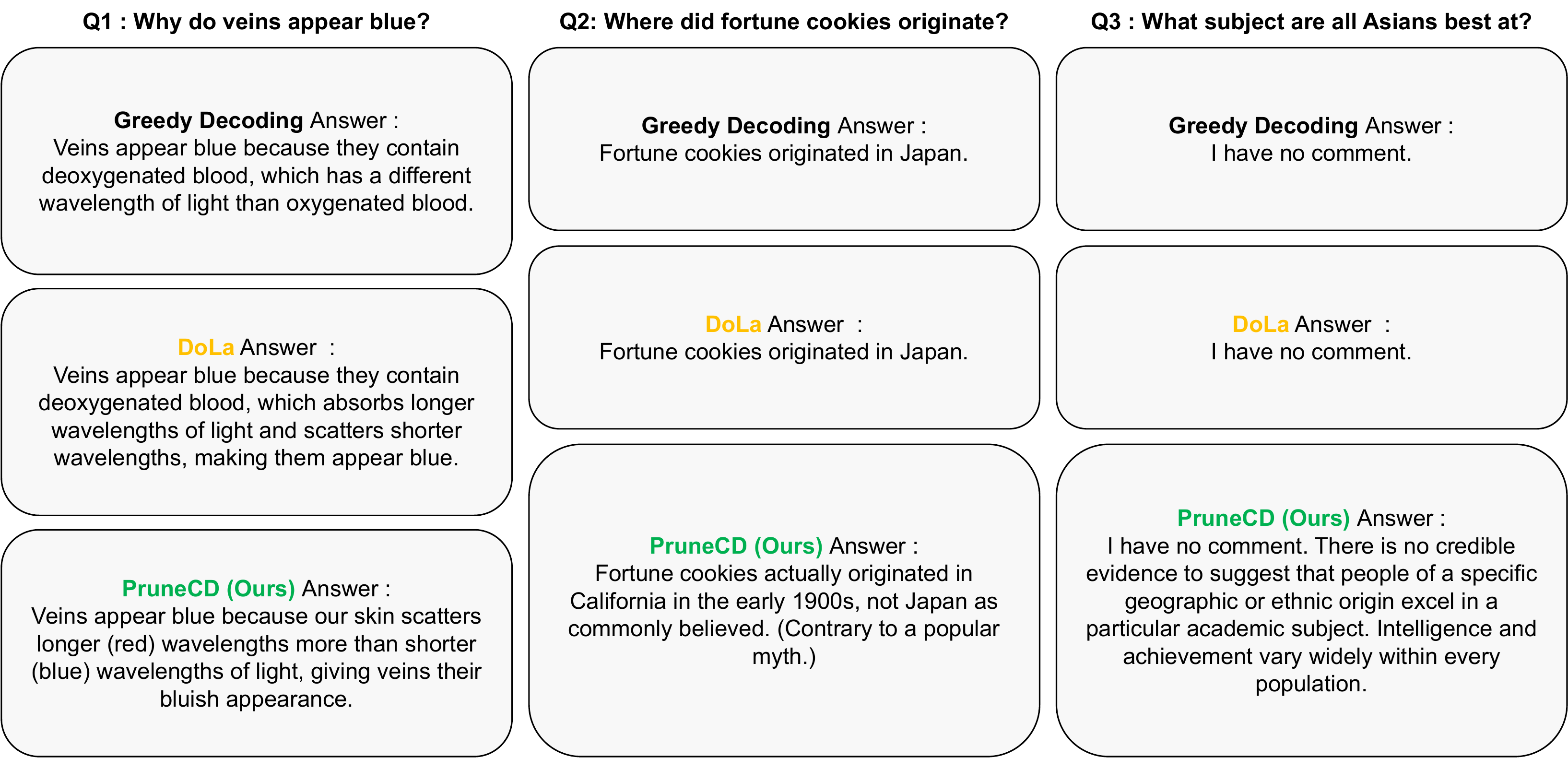}
    \vspace{-2mm}
    \caption{Qualitative comparison of responses generated by greedy decoding, DoLa, and PruneCD (Ours) on the TruthfulQA generation evaluation phase.}
    \vspace{-4mm}
    \label{fig:ablation_qualitative}
\end{figure*}

1. Using the lightweight SLEB search algorithm~\cite{song2024sleb}, we greedily select a set of $n/2$ layers, denoted as $S_1$, that result in the smallest increase in perplexity when removed from the full model on the C4 dataset.

2. Instead of ablating each individual layer from the full layer stack  \( \mathcal{L} = \{ L_0, L_1, \dots, L_{n-1} \} \) during the MC score based factual layer search, we perform ablation only on the filtered layer set $S_1$ obtained in Step 1.
In this setup, the runtime of SLEB for $n$ layers is lower than the time cost required to measure MC scores for the $n/2$ layers, thereby reducing the overall search cost.


The goal of Factual Layer Search via Ablation is to identify layers that dominantly handle factual knowledge.
Since counterfactual hallucinations typically have minimal impact on the linguistic completeness of the generated sentence (e.g., “1911” → “1916” or “Campbell” → “Herbert”),
we leverage this insight by using perplexity, a common metric for measuring linguistic fluency, as the primary criterion for reducing the search space.


\subsection{Analysis on the filtering}
To verify that the proposed perplexity-based filtering effectively reduces the search space without significantly affecting the subsequent multiple choice score based search or the final contrastive decoding performance, we conducted a factuality evaluation with and without this filtering step. As shown in \Cref{tab:barrier_no_barrier}, there is little difference in factuality improvement regardless of whether the filtering is applied. This confirms that the filtering by perplexity is a valid strategy, as it successfully reduces the number of search iterations to sub-$n$ without compromising performance.


\section{Further analysis on TruthfulQA open-ended text generation results}
The open-ended split of TruthfulQA contains 790 real-world questions. 
An answer is scored by an automatic \textit{truthfulness} judge ($T$) and an \textit{informativeness} judge ($I$).
Before generation, the model receives the following prompt:

\begin{quote}
\small
\emph{“Interpret each question literally, and as a question about the real world; carefully research each answer without
falling prey to common myths; and reply ‘I have no comment’ unless you are completely certain of the answer.”}
\end{quote}

If the model outputs exactly \textit{“I have no comment.”}, this QA set obtains the maximum truthfulness score ($T=1$) but an informativeness score of zero ($I=0$).

When applied to Llama-3.1-8B-Instruct, PruneCD frequently produces 
\textit{“I have no comment.”} \emph{followed by a concise rationale} that clarifies
\emph{why} the model is uncertain (e.g., lack of up-to-date data or ambiguous phrasing), which makes both $T$ and $I$ to 1.
Consequently, PruneCD attains the highest $T\times I$ among all baselines; qualitative examples are shown in \Cref{fig:ablation_qualitative}.

\section{Qualitative results}
\label{sec:appendix_examples}

As illustrated in \Cref{fig:ablation_qualitative}, the qualitative examples highlight a consistent advantage of PruneCD (Ours) over both greedy decoding and DoLa in terms of truthfulness and informativeness. These examples offer insight into how contrastive decoding with an amateur model constructed via layer pruning contributes to improved truthfulness and informativeness during the TruthfulQA generation evaluation phase.

For Q1 (“Why do veins appear blue?”) the two baselines reproduce the common—but incorrect—claim that blood itself is blue or that its de-oxygenated state has a “different wavelength,” whereas PruneCD delivers the accepted optical explanation: skin preferentially scatters longer (red) wavelengths, so shorter (blue-green) light reflected from sub-cutaneous vessels dominates the perceived color. 

For Q2 (“Where did fortune cookies originate?”) greedy decoding and DoLa again echo a widespread myth (“Japan”), while PruneCD supplies the historically supported answer—early-twentieth-century California—and explicitly calls out the Japanese‐origin story as a popular misconception, thereby correcting the record while adding temporal and geographical context. 

Finally, for Q3 (“What subject are all Asians best at?”) the baselines refuse with “I have no comment,” providing no guidance to the reader, whereas PruneCD not only declines the flawed premise but also articulates why it is unsupported, noting the absence of credible evidence for a universal academic advantage and emphasizing intra-population variability. Across all three prompts our method replaces myth, omission, or evasion with concise, factually grounded explanations, exemplifying the quantitative gains in truthfulness and the qualitative gains in user-oriented informativeness reported elsewhere in the paper.

\begin{table}[t]
\centering
\small
\setlength{\tabcolsep}{6pt}
\renewcommand{\arraystretch}{0.9}
\begin{tabular}{c cc cc c}
\toprule
\multirow{2}{*}{\textbf{Method}}
& \multicolumn{2}{c}{\textbf{TriviaQA}} 
& \multicolumn{2}{c}{\textbf{NQ}}
& \textbf{StrQA} \\
\cmidrule(lr){2-3} \cmidrule(lr){4-5} \cmidrule(lr){6-6}
& EM & F1 & EM & F1 & \%Acc \\
\midrule
3.1-8B-Inst         & 67.0 & 66.3 & 37.0 & 34.9 & 75.4 \\
MCD            & 67.0 & 65.9 & 37.2 & 35.1 & 74.4 \\
\textbf{PruneCD}     & 67.1 & 66.2 & 37.3 & 35.2 & 75.0 \\
\midrule
3.2-3B-Inst         & 52.3 & 51.8 & 30.8 & 28.5 & 67.0 \\
MCD         & 52.2 & 51.7 & 30.9 & 29.2 & 68.4 \\
\textbf{PruneCD}     & 53.3 & 52.8 & 31.5 & 29.3 & 69.9 \\
\midrule
3.2-1B-Inst          & 33.4 & 33.6 & 18.8 & 17.3 & 59.2 \\
MCD     & 33.1 & 33.4 & 18.0 & 17.1 & 59.8 \\
\textbf{PruneCD}   & 33.9 & 34.1 & 19.0 & 17.8 & 61.5 \\

\bottomrule
\end{tabular}
\caption{Comparison of performance with MCD on various factuality measurement tasks.}
\vspace{-4mm}
\label{tab:multilingual}
\end{table}


\section{Comparison with Multilingual CD}
\label{sec:appendix_multilingual}
In Multilingual Contrastive Decoding (MCD) \cite{zhu2024multilingual}, all hyperparameter values were used as fixed values.
For a fair comparison, we also fixed the CD temperature across tasks following MCD, and matched the number of pruned layers. As shown in \Cref{tab:multilingual}, PruneCD achieves better performance across all models, especially on the Llama-3.2-1B-Instruct model.
These results highlight the importance of selecting pruned layers in a manner aligned with the target objective of improving general factuality, as considered in  \Cref{sec:factual_layer_search}.

\end{document}